# Recognition and Processing of NATOM Based on Decoupling Features and Classifiers

Yinhui, Luo

The Computer School, Civil Aviation Flight University of China

Yipeng, Deng

The Computer School, Civil Aviation Flight University of China, 931632121@qq.com

**ABSTRACT**

In this paper we show how to process the NOTAM (Notice to Airmen) data of the field in civil aviation. The main research contents are as follows:

- Data preprocessing: For the original data of the NOTAM, there is a mixture of Chinese and English, and the structure is poor. The original data is cleaned, the Chinese data and the English data are processed separately, word segmentation is completed, and stopping-words are removed. Using Glove word vector methods to represent the data for using a custom mapping vocabulary.

- Decoupling features and classifiers: In order to improve the ability of the text classification model to recognize minority samples, the overall model training process is decoupled from the perspective of the algorithm as a whole, divided into two stages of feature learning and classifier learning. The weights of the feature learning stage and the classifier learning stage adopt different strategies to overcome the influence of the head data and tail data of the imbalanced data set on the classification model.

Experiments have proved that the use of decoupling features and classifier methods based on the neural network classification model can complete text multi-classification tasks in the field of civil aviation, and at the same time can improve the recognition accuracy of the minority samples in the data set.

## 1. INTRODUCTION

In the field of civil aviation, the NOTAM contains important key information for flight crews, ground control personnel, and maintenance logistics support personnel. However, the current manual processing methods have increased repetitive human labor and caused more human errors. the NOTAM data has the characteristics of mixed Chinese and English and uneven distribution. According to the characteristics of the NOTAM data set, this paper has conducted some research and practice on multiple types of text classification. Complete tasks such as sorting and processing the notices of navigation, assisting the crew, and providing assistance for control, dispatch and flight crews. This can effectively reduce the burden on the crew, minimize the impact of human factors on flight work, ensure flight safety, and lay the foundation for the digitalization, informatization, and automation of civil aviation.

## 2. REALTED WORK

To solve the issue about unbalanced dataset, researchers have conducted some research. The current mainstream methods are as follows: The first is Re-Sampling, which includes oversampling of minority sample[1] and under-sampling of majority samples[2]. However, oversampling tends to cause the model to learn too many features that should not be learned, thereby reducing the generalization ability of the model, and the robustness is not strong, while under-sampling often leads to the model not learning enough features, and there is underfitting.

The second is data synthesis (Synthetic Samples). As far as text classification tasks are concerned, by generating 'similar' data from minority samples, the classic method is to synthesize the minority class algorithm (SMOTE, Synthetic Minority Over-Sampling Technique) algorithm[3], By selecting arbitrary minority samples, using k-nearest neighbor algorithms to synthesize similar data. This will not lead to overfitting of the model, nor will it make the model unable to learn enough features, which effectively enhances the generalization ability of the model. The third type is Re-Weighting. The weights of different categories are not the same, and the weights can be changed. There are many ways to do weighting.

The fourth type is Transfer Learning, which is more common in deep learning. By separately modeling the majority class samples and the minority class samples, in general, under the same data set, all class samples have a certain similarity. The characteristics of the degree. Then the knowledge or features can be transferred to the minority samples through the learned features of the majority samples. Liu[4] proposed a new algorithm for Open Long Tail Recognition (OLTR) based on dynamic meta-embedding. The main work is to embed visual information from the head and tail to enhance the robustness of tail recognition. At the same time improve the sensitivity of open recognition. Yin[5] proposed a center-based feature transfer framework to expand the feature space of under-represented subjects in a sufficiently diverse routine.

The fifth type is metric learning. Essentially, it allows the model to learn a better text representation, so that the model has a more appropriate degree of discrimination for the decision boundary of the minority samples. Among them, Zhang[6] proposed a new loss function called range loss in order to effectively use the entire long-tail data in the training process. The scope loss aims to reduce the overall intra-personal disparity, while at the same time expanding the inter-personal disparity in a mini-batch when faced with extremely unbalanced data. Huang[7] showed that by forcing deep networks to maintain profit rates between clusters and classes, more discriminatory deep representation can be learned. This stricter constraint effectively reduces the class imbalance inherent in the local data community.

The sixth is meta learning, which processes the head and tail data differently, and can adaptively learn to re-weight[8]. The seventh is the decoupling representation and classifier[9] (decoupling representation and classifier). The text classification process is divided into two stages. The first stage learns its characteristics, and the sampling is balanced in the classifier learning stage. The current result is the best. Zhou[10] proposed a unified Bi-directional Branch Network (BBN, Bi-directional Branch Network), which handles both representative learning and classifier learning. Each branch does perform its own duties separately. In particular, the BBN model is further equipped with a novel cumulative learning strategy, which aims to learn the general pattern first, and then gradually focus on the tail data.

## 3. METHODS

For the unbalanced NOTAM data, this article first uses the neural network method to represent the original data as a low-dimensional, dense word vector matrix. Then use an algorithm optimization method to split the neural network training model into two stages, feature extraction learning and classifier learning. In the two stages, different strategy processing is adopted to artificially control the sample sampling feature weight of the model to enhance. The classification model's ability to recognize and process minority samples. The overall flow chart of the text multi-classification algorithm research and implementation is shown in Fig. 1.

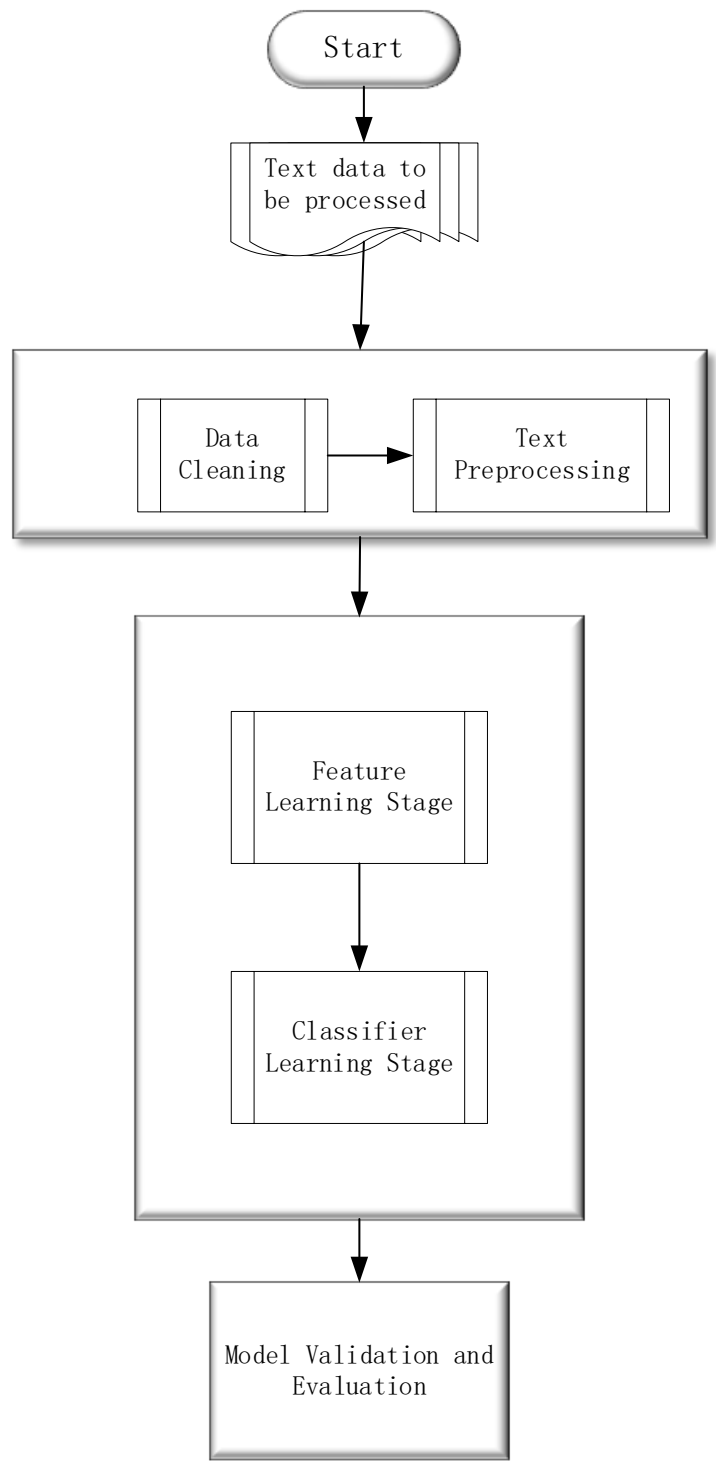

Fig. 1. The Overall Process of the Text Classification of the NOTAM

### 3.1. Data Set Introduction

The NOTAM is a general term for all the information about flight operations in the civil aviation field. It is issued by official agencies. It is an information reminder to relevant personnel performing flight operations in designated airspace, including possible hazards on the route and on the ground. Flight information area data Update, as well as aircraft take-off, approach to the airport related information, etc.

As the types of NOTAM are relatively fixed and suitable for computer processing, the use of artificial intelligence methods to automatically complete the notice text classification of NOTAM can greatly reduce the workload of the staff and reduce errors. At present, there are many characteristics NOTAM. The Chinese and English are mixed, the structure is poor. Part of the original data is shown in Table 1.

TABLE 1. the part of the original data

| content | label |
| --- | --- |
| 以和田机场为中心半径100KM 范围内,高度在6600M( 含) 至9800M( 含) 之间禁航. | RTLP |
| DME 05 'IWF' CH40X 仅供测试，不可使用，因校飞. | IDCT |
| UNMANNED AIRCRAFT ACTIVITY WILL TAKE PLACE RADIUS 1000M CENTERED ON 410301N0083917W. | WULW |

There are 113 types of data in the original data set, and the total number of data is 760,892. The smallest label is "PICN", which has 1002 samples. The most label is "MXLC", with 87078 samples. The number of categories with the largest sample data is more than 80 times that of the smallest sample. The label data distribution is shown in Fig. 2.

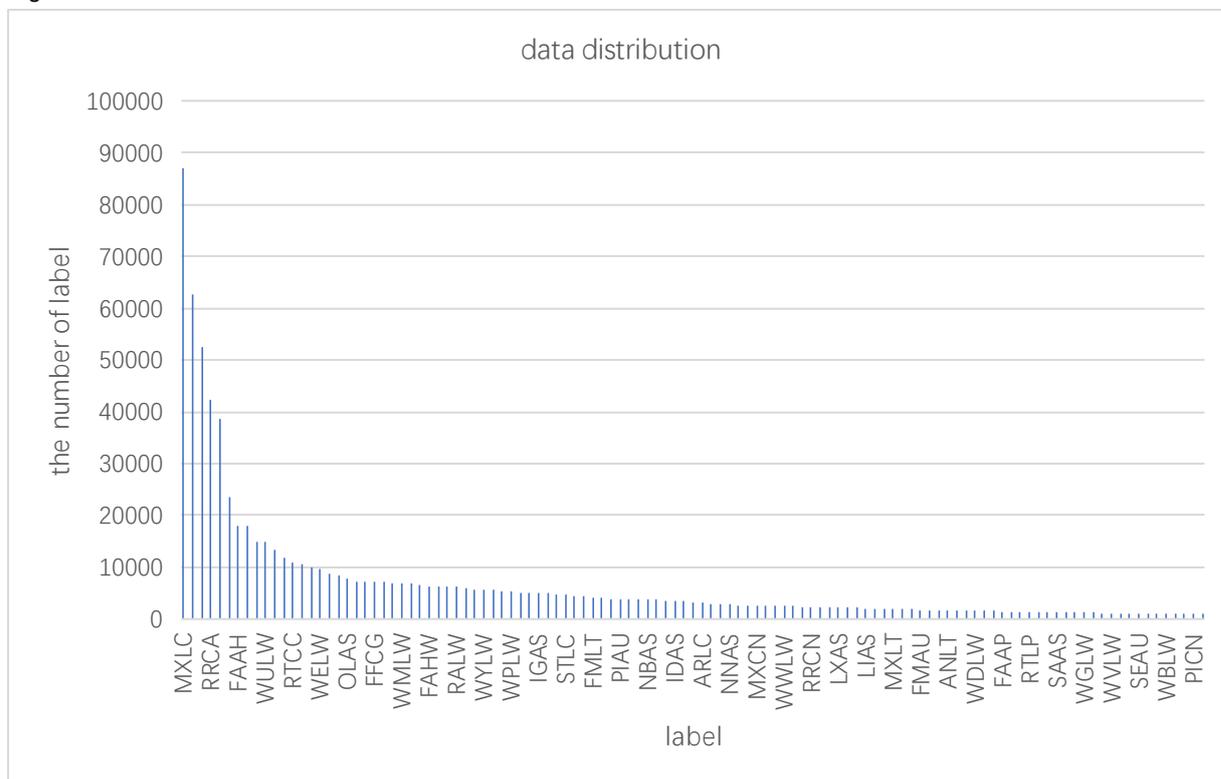

Fig. 2. the data distribution

### 3.2. The Feature Learning

In the feature learning stage, the framework selection strategy in this article has the following four types. The first is standard sampling, that is, the instance sampling method. It does not perform operations such as up-sampling and down-sampling on the data. The probability of each category being sampled It is related to the proportion of this category in the data set, and its probability can be expressed as shown in formula 1.

$$P_i^{IBS} = \frac{m_i}{\sum_{j=1}^{s} m_j} \tag{1}$$

Where S is the number of categories, $P_i$ is the sampling probability of the i-th category in the training set, and $m_i$ is the number of the i-th sample in the training set.

The second is the CBS (class-balanced sampling) method, where each type of data has the same probability of being sampled. This article is no longer based on instance sampling, to ensure that in the feature learning stage, each category can be learned with the same probability. The probability can be expressed as shown in formula 2.

$$P_i^{CBS} = \frac{1}{S} \tag{2}$$

Generally speaking, the oversampling method can be directly used for the minority samples to make the number of samples consistent with the majority of samples. First select a category from the category set, and then sample the data of this category with equal probability. The algorithm flow is shown in Table 2.

TABLE 2. Class balance sampling algorithm flow

| CBS class balance sampling algorithm |
| --- |
| ● Write raw data |
| ● The original data is grouped by category, that is, the same category is entered into the same list, and the original data is converted into the form of list (list()), in which the internal data is sorted randomly |
| ● Sampling the original data with equal probability according to formula (2) |
| ● Loop sampling completed |

The third is square root sampling (SRS), which is also a variant of class balance sampling. In class balance sampling method, oversampling of minority classes can easily lead to overfitting. By sampling square root sampling, the model can be suppressed to a certain extent. Fit, as shown in formula 3.

$$P_i^{SRS} = \frac{\sqrt{m_i}}{\sum_{j=1}^{s} \sqrt{m_j}} \tag{3}$$

In formula 3.28, S is the number of categories, $P_i$ is the sampling probability of the i-th category in the training set, and $m_i$ is the number of the i-th sample in the training set. The sampling process of the square root algorithm is similar to the class balance algorithm.

The fourth is the introduction of the gradual balanced sampling strategy (PBS), which is essentially a hybrid sampling method that combines the instance-balanced sampling method with the category-balanced method. The method uses the example balance sampling method in the early stage. After the sampling iteration period gradually increases, the sampling method gradually switches to the category balance algorithm. The formula is shown in formula 4.

$$P_i^{PBS} = \frac{t}{T} P_i^{CBS} + \left(1 - \frac{t}{T}\right) P_i^{IBS} \tag{4}$$

T is the total iteration period, and t is the t-th round of the iteration period.

The PBS sampling algorithm is shown in Table 3.

TABLE 3. PBS sampling algorithm flow

| **PBS class balance sampling algorithm** |
|---|
| ● Write raw data |
| ● The original data is grouped by category, that is, the same category is entered into the same list, and the original data is converted into the form of list (list()), in which the internal data is sorted randomly |
| ● Determine the conversion frequency of the sampling model according to the iteration period. The expression is numpy.linspace(0,1,epochs) where epochs is the model training iteration period |
| ● According to formula (4) to sample the original data cyclically |
| ● Loop sampling completed |

### 3.3. The Classifier Learning

In this section, a new method of balancing the weights of the classifier is selected to realize the method of identifying and processing head data and tail data, so as to adjust the classifier decision boundary of clear head data and tail data. Experiments show that there is no need for any Other retraining, the classifier still has a good recognition and classification effect. After the first stage of feature learning and sampling of the classifier is completed, it enters the second round of classifier learning stage. In order to make the model perform better on minority samples, a new training method needs to be adopted in the classifier stage.

1) Classifier Re-Training method (CRT)

Re-initialize the classifier weights randomly and use the class balance method for training. In the feature extraction part, the weights need to be fixed during training. While keeping the parameters fixed, the class balance sampling method is used to randomly process the minority samples, including reinitialization and optimization of the weight w and bias b of the classifier. The classifier retraining algorithm flow is shown in Table 4.

TABLE 4. Classifier retraining algorithm flow

| Classifier retraining algorithm flow |
|---|
| ● Write raw data |
| ● Sampling the original data in the feature learning stage, completing feature learning, training and saving the training parameters |
| ● Import the saved training parameters in the classifier learning stage, and then initialize the classifier weights |
| ● The weight parameters of the fixed feature representation stage, the class balance method is used to resample the data to retrain the classifier, and the weight parameters w and bias b are updated. |
| ● The training is completed and the results are evaluated |

2) Nearest neighbor class balance classifier (NCM)

The core idea is to first calculate the average feature representation of each category on the training set, and then use Euclidean distance or Mahalanobis distance to assign data close to the mean to the category.
As shown in formula 5.

$$\bar{y} = argmin_{y \in \{1...Y\}} p(x, \mu_y) \qquad (5)$$

Where Y represents the number of categories and the average value of the categories $\mu_y = \frac{1}{N_y} \sum_{i:y_i=y}^{N_y} x_i$. $N_y$ is the number of category y. The key to the success of the nearest neighbor class balancing classifier lies in the measurement of distance. The classification can be effectively completed when the Mahalanobis distance between the categories is appropriate. Learn Euclidean distance by maximizing the log likelihood function. See formula 6.

$$\mathcal{L} = \frac{1}{N}\sum_{i=1}^{N} \ln P(y_i|x_i) \tag{6}$$

$$p(y|x) \propto e^{-0.5 p_{xy}^w} \tag{7}$$

$$p_{xy}^w = (x - \mu_y)^T W^T W (x - \mu_y) \tag{8}$$

In formula 6, formula 7, formula 8, $x \in \mathbb{R}^D$, $W \in \mathbb{R}^{m \times D}$, m≤D is the fixed dimension of the metric space. The final NCM classifier can be considered as a multi-class Softmax classifier with limited deviation terms and weight vectors, and the constraint conditions of the weight vectors are shown in formula 9.

$$w_y = W^T W \mu_y \tag{9}$$

In the training process, NCM does not rely on a fixed representation and the metric W of learning Mahalanobis distance, but learns the representation of deep learning φ(·). It is assumed that the high-order nonlinear nature of the deep learning representation eliminates the need for the metric W, and the Euclidean distance between deep learning representations can be used. As shown in formula 10.

$$p_{xy}^w = (\varphi(x) - \mu_y^\varphi)^T (\varphi(x) - \mu_y^\varphi) \tag{10}$$

$$\mu_y^\varphi = \frac{1}{N} \sum_{i:y_i=y}^{Y} \varphi(x_i) \tag{11}$$

In formula 11, $\varphi(x_i)$ is the deep representation of data x, and the result is obtained after every training calculation is optimized.

When updating the mean, directly estimate the mean. As shown in formula 12.

$$\mu_{y_i}^\varphi = \frac{n_{y_i}}{n_{y_i}+1} \mu_{y_i}^\varphi + \frac{1}{n_{y_i}+1} \varphi(x_i) \tag{12}$$

In formula 12, $n_{y_i}$ is the number of samples in the current category $yi$, each batch of learning uses the current sample to update the mean.

Use the attenuation scheme to update after each training. As shown in formula 13.

$$\mu_{y_i}^\varphi = \alpha \mu_{y_i}^\varphi + (1-\alpha)\varphi(x_i) \tag{13}$$

Use the mean of this batch during the training process instead of using a single data sample $\varphi(x_i)$

The training process is shown in Table 5.

TABLE 5. NCM sampling classifier training process

| NCM classifier training process |
|---|
| ● Write original data |
| ● Sampling the original data in the feature learning stage, completing feature learning, and training to save the training parameters |
| ● Import the saved training parameters in the learning stage of the classifier, and calculate the average feature representation of each category |
| ● Calculate the cosine similarity or Mahalanobis distance between categories, use the KNN clustering algorithm to search, and adjust the decision boundary of the classifier. Optimize the recognition ability of head and tail data. |
| ● The training is completed and the results are evaluated. |

## 4. EXPERIMENTS

In this article, the model is based on the Paddle framework, the model uses TextCNN as the backbone network, and a fully connected layer as the classifier. In the feature learning stage, save the TextCNN backbone network parameters after learning with different sampling strategies. In the classifier learning stage, load the saved TextCNN backbone network parameters, and freeze the TextCNN backbone network parameters during training, and only train and update the fully connected layer parameters.

we use the TextCNN neural network as the Baseline model. The batch size is 64, and Adam is used to train the model. The learning rate of the first 10 iterations during training is 5e-5, and the learning rate of the next 5 iterations is updated to 5e-6. Fig. 3 shows the model training results.

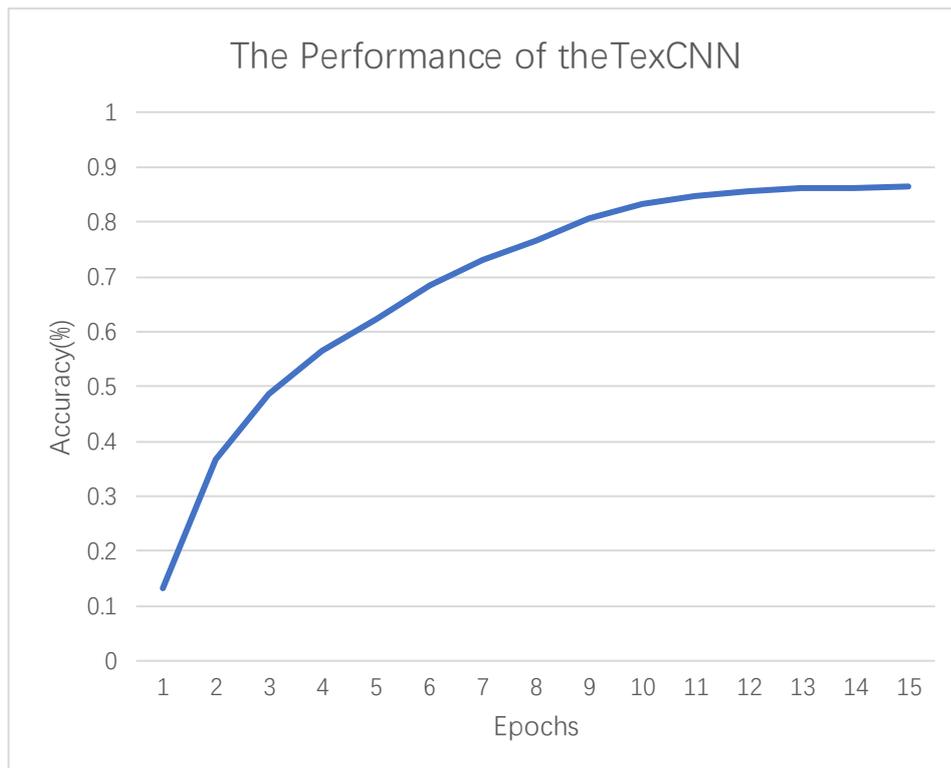

Fig. 3. the Performance of TextCNN

After 15 epochs of training, the model converged in about 13th epoch, and the accuracy rate finally reached about 86.4%. The running time of the TextCNN model is 1 hour and 22 minutes.

In the experiment process of this article, the strategies adopted in the two stages are different. There are 8 methods in this framework. Note: The training results under this framework are all final results, which are the results after training in the two stages. Among them are the IBS algorithm combined with the CRT classifier, the IBS algorithm combined with the NCM classifier, the CBS algorithm combined with the CRT classifier, and the CBS algorithm combined with the NCM classifier. The SRS algorithm combined with CRT classifier, the SRS algorithm combined with NCM classifier, the PBS algorithm combined with CRT classifier, the PBS algorithm combined with NCM classifier. The effect diagram is shown in Table 6.

TABLE 6. The PERFORMANCE of Decoupling feature and classifier algorithm model

|  | CRT classifier | NCM classifier |
| --- | --- | --- |
| IBS algorithm | 0.7221 | 0.73454 |
| CBS algorithm | 0.6541 | 0.54548 |
| SRS algorithm | 0.6654 | 0.6712 |
| PBS algorithm | 0.8653 | 0.84545 |

Compared with the Baseline model, the overall training effect is actually not particularly good. The combination of the above sampling algorithm strategy and the classifier weight strategy is even slightly less effective. In particular, the effect of the category balancing algorithm and the square root sampling algorithm is not good enough. Compared with the benchmark model, the overall effect is decreased, only when the stepwise balance algorithm is combined with the classifier, the effect is close to the benchmark method.

In order to verify that the data enhancement algorithm can effectively enhance the model's recognition accuracy for minority samples, we use 'much' to represent the recognition accuracy of the head data MXLC, 'medium' to represent the recognition accuracy of the median data OBCN, and 'less' to represent Recognition accuracy rate of tail data PICN.

TABLE 7. The PERFORMANCE of decoupling features and classifier algorithms in different categories of data

| 3. | much | medium | less |
| --- | --- | --- | --- |
| baseline | 0.921 | 0.344 | 0.171 |
| the IBS algorithm combined with the CRT classifier | 0.742 | 0.563 | 0.321 |
| the IBS algorithm combined with the NCM classifier | 0.721 | 0.541 | 0.336 |
| the CBS algorithm combined with the CRT classifier | 0.684 | 0.37 | 0.181 |
| the CBS algorithm combined with the NCM classifier | 0.692 | 0.351 | 0.176 |
| the SRS algorithm combined with CRT classifier | 0.729 | 0.391 | 0.226 |
| the SRS algorithm combined with NCM classifier | 0.718 | 0.426 | 0.21 |

When the model is required to identify the head data in the original data, that is, most types of data, the baseline model can achieve good results, with an accuracy rate of about 92.1%. When the model is required to identify the median sample in the data set, based on the baseline model combined with the IBS algorithm and CRT classifier strategy, compared with the baseline model, the recognition accuracy of the median data sample is improved by 21.9%, reaching 56.3%. If the model is needed to strengthen the recognition accuracy of the minority samples in the data set, the IBS algorithm and NCM classifier are combined on the basis of the baseline model. Compared with the baseline model, the accuracy is improved by 16.5%. Through the combination of the above strategies, the model has improved the recognition accuracy of minority samples and median samples. But it is obvious that this is at the cost of sacrificing the ability to recognize and process the head data.

## 5. CONCLUSIONS

In this paper, the text multi-classification task in the imbalanced data set has been researched and realized. However, the work done in this article is limited and restricted under certain conditions. In the future work, there are still many places worth exploring and improving.

There are 1434 categories in the original data set. In this article, for the convenience of processing and optimization of the results, when the amount of data of the category is less than 1000, it will be cleaned out. There are always 113 categories in this article. All of the 1321 categories have been cleaned out. How to identify the remaining 1321 types of "tail" data is a direction worth studying in the future.

In the decoupling feature and classifier learning method, although the model has a certain improvement in the recognition and processing of minority samples, it is at the expense of head data recognition and processing capabilities. In theory, the decoupling feature and classifier framework can not only enhance the model's ability to recognize minority samples, but also does not sacrifice the recognition and processing ability of head data. This proves that the training parameters of the model, the selection of the sampling algorithm, and the selection of the classifier weight algorithm all have great room for improvement.

## REFERNCES